%% file: sample-sigconf.tex
\documentclass[sigconf]{acmart}

\AtBeginDocument{%
  }

\begin{document}

\title{Medical Data Augmentation via ChatGPT: A Case Study on Medication Identification and Medication Event Classification}

%
\author{Shouvon Sarker, Lijun Qian, Xishuang Dong}
\affiliation{%
  \institution{CREDIT, Prairie View A\&M University}
  \streetaddress{100 University Dr}
  \city{Prairie View}
  \country{US}}
\email{ssarker3@pvamu.edu, liqian@pvamu.edu, xidong@pvamu.edu}


%
%
%
%


\begin{abstract}

The identification of key factors such as medications, diseases, and relationships within electronic health records and clinical notes has a wide range of applications in the clinical field. In the N2C2 2022 competitions, various tasks were presented to promote the identification of key factors in electronic health records (EHRs) using the Contextualized Medication Event Dataset (CMED). Pretrained large language models (LLMs) demonstrated exceptional performance in these tasks. This study aims to explore the utilization of LLMs, specifically ChatGPT, for data augmentation to overcome the limited availability of annotated data for identifying the key factors in EHRs. Additionally, different pre-trained BERT models, initially trained on extensive datasets like Wikipedia and MIMIC, were employed to develop models for identifying these key variables in EHRs through fine-tuning on augmented datasets. The experimental results of two EHR analysis tasks, namely medication identification and medication event classification, indicate that data augmentation based on ChatGPT proves beneficial in improving performance for both medication identification and medication event classification.


  \end{abstract}

%



\maketitle

\section{Introduction}
\label{sec1}
\input{Introduction}

\section{Methodology}
\label{sec4}
\input{Method}

\section{Experiment}
\label{sec5}
\input{Experiment}
\section{Related Work}
\label{sec3}

\input{Relatedwork}


\section{Conclusion}
\label{sec7}
\input{Conclusion}
\section{Acknowledgments}
The material is based upon work supported by NASA under award number 80NSSC22KM0052.
Any opinions, findings, and conclusions or recommendations expressed in this material are those of the
author(s) and do not necessarily reflect the views of NASA.

\bibliographystyle{ACM-Reference-Format}
\bibliography{references}

\end{document}

%% file: Introduction.tex

\begin{figure*} [ht]
 	\centering	\includegraphics[width=0.9\linewidth]{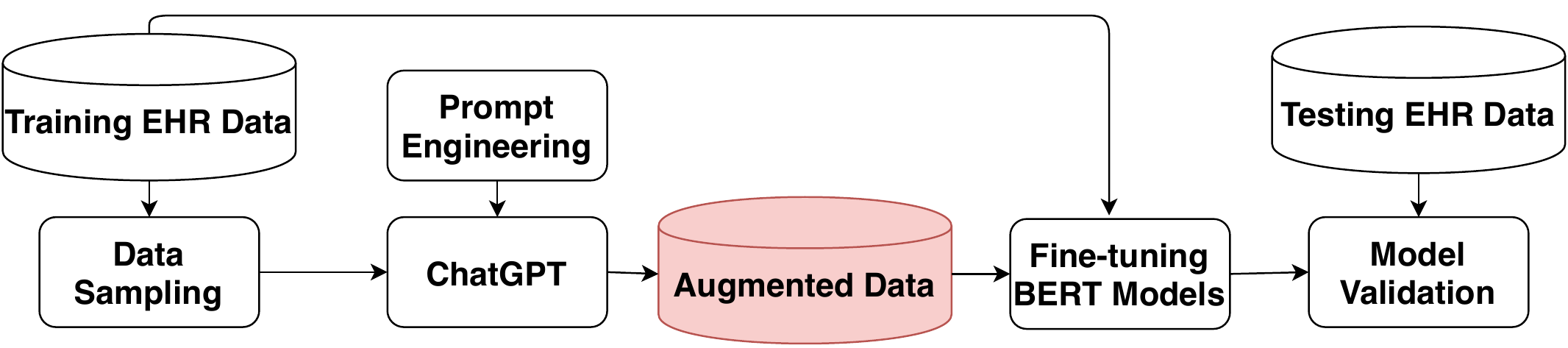}
	\caption{Pipeline of Data Augmentation Based on ChatGPT for Enhancing EHR Data Analysis.}
	\label{Fig_data_aug}
\end{figure*}

Electronic Health Records (EHRs) typically consist of unstructured text containing information about a patient's medical history, health conditions, discharge papers, and clinical notes. The unstructured nature of these texts poses a significant challenge for processing using state-of-the-art language model representations~\cite{GoodBengCour2016}. Additionally, the scarcity of clinical datasets due to privacy and patient security concerns further complicates the training of deep learning models. To encourage advancements in data analytics on EHRs, the N2C2 2022 competitions have invited teams to participate in various tasks aimed at identifying key factors such as medications, diseases, and relationships within the Contextualized Medication Event Dataset (CMED)~\cite{cmed}.

Over the past few years, there has been a significant breakthrough in natural language processing (NLP) tasks with the introduction of pretrained large language models (LLMs) such as BERT. These LLMs are transformer-based architectures that undergo unsupervised training on extensive text data to comprehend the intricate features and patterns of human language. During the pre-training phase, LLMs acquire the ability to predict the subsequent word in a sentence by leveraging the context provided by preceding words, enabling them to capture intricate semantic features and patterns~\cite{Radford2018ImprovingLU}. Following pre-training, LLMs can be fine-tuned for specific downstream tasks by training them on task-specific labeled datasets.

In this study, a novel approach for data augmentation in EHR analysis is proposed, utilizing prompt engineering with ChatGPT to leverage the capabilities of LLMs. The process involves three key steps: 1) Data augmentation through ChatGPT: The original sentences are rephrased using ChatGPT, thereby generating augmented data; 2) Fine-tuning pre-trained BERT models: The pre-trained BERT models are fine-tuned using both the original CMED training dataset and the augmented dataset obtained from ChatGPT. 3) Application of fine-tuned BERT models: The fine-tuned BERT models are then applied to two tasks, namely medication identification and medication event classification, using the CMED test data. The experimental results demonstrate the effectiveness of the proposed data augmentation approach, particularly in enhancing the performance of medication event classification. Notably, the augmentation technique improves the Recall metric for medication event classification.

In summary, contributions of this work are summarized as bellow.

\begin{itemize}
\item We explore the utilization of ChatGPT for implementing data augmentation in EHR data analysis through prompt engineering.

\item  We utilize different variants of BERT models to evaluate the effectiveness of augmented data for EHR data analysis. Specifically, we assess the performance on two tasks: medication identification and medication event classification. By examining the results from various perspectives, we validate the impact of augmented data in enhancing the performance of these tasks in EHR data analysis.

\end{itemize}


The remaining sections of this paper are structured as follows. Section 2 presents a detailed explanation of the proposed method. Section 3 provides a comprehensive discussion and presentation of the experimental results. Section 4 delves into an in-depth exploration of related work. Section 5 concludes the paper by summarizing the key findings and drawing overall conclusions.

%% file: Method.tex


Large language models (LLMs) have found wide-ranging applications in natural language processing tasks, such as question answering, text generation, and language translation~\cite{zhao2023survey}. In this study, the focus lies on augmenting the data using ChatGPT. Subsequently, pre-trained BERT models are trained on the original training data as well as the training data combined with the augmented data. The results demonstrate an improvement in performance when the models are trained using the augmented data.

\subsection{GPT}

The Generative Pre-trained Transformer (GPT) is an advanced language model developed by OpenAI that has demonstrated impressive performance across various natural language processing tasks. Built upon the Transformer architecture, GPT employs self-attention mechanisms to capture the relationships between different tokens within a sequence. Through unsupervised learning via masked language modeling, GPT is pre-trained on an extensive corpus of text data~\cite{Radford2018ImprovingLU}. One of the notable strengths of GPT is its capacity to generate text that exhibits human-like qualities, including coherence and stylistic consistency. This text generation process involves sampling from a probability distribution over the language's vocabulary~\cite{4}. In this study, we utilize ChatGPT, a variant of GPT, to generate additional training data samples, thereby enhancing diversity and introducing linguistic variations to the dataset.

%
%

%

\subsection{BERT Model}

In this study, we employed BERT to construct multiple baselines (transformer). During training, BERT receives a pair of sequences and attempts to predict whether the second sequence in the pair originates from the same document. For training purposes, 50\% of the sentences are sourced from the same documents, while the other 50\% are randomly selected from different documents to serve as the second sequence (word).During prediction, it is assumed that the second sequence will be distinct from the first sequence~\cite{bert}. To differentiate between the two sentences, BERT conventionally inserts a [CLS] token at the beginning of the sequence and a [SEP] token at the end of the sequence. These [CLS] and [SEP] tokens aid BERT in discerning the sentences~\cite{bert}. However, BERT's performance may not be optimal when dealing with clinical and biomedical texts, as the vocabulary representation and distribution in these domains can differ significantly from those in general domains. To address this limitation, BERT requires pre-training on clinical data~\cite{bert, overview}.

\subsection{Prompt Engineering}
In the field of natural language processing, the design and modification of prompts to elicit specific responses from LLMs are crucial. LLMs like ChatGPT, which excel at generating text resembling human speech, have garnered significant research interest in recent years~\cite{white2023prompt}. In this research, the objective of prompt engineering is to create prompts that allow the language model to achieve optimal performance on the rephrasing task, generating data while maintaining the underlying conditions of patients and their medications. This process involves defining an objective function, specifically focusing on medication events, that captures the desired behavior of the model. Pertinent prompts are carefully chosen, and the language model's parameters are adjusted to optimize the objective function.


\subsection{Enhancing EHR Data Analysis via Data Augmentation Based on ChatGPT}

Data augmentation aims to increase the size and diversity of the training dataset, which helped improving the performance and generalization of the pre-trained models. Figure~\ref{Fig_data_aug} presents the pipeline of enhancing EHR data Analysis via data augmentation based on ChatGPT.


This pipeline begins by randomly sampling training samples to serve as inputs for generating data using ChatGPT. Simultaneously, tailored prompts are created through prompt engineering to cater to the specific tasks of EHR data analysis. One concern with augmenting data using ChatGPT is that the generated text might lack diversity or contain biases and errors present in the training data. To tackle this, we meticulously choose prompts and closely monitor the quality of the generated text data, ensuring it doesn't introduce bias towards a specific class of medication events. Once the augmented data is acquired, it is combined with the training EHR data to perform fine-tuning of BERT models. Finally, the fine-tuned BERT models undergo validation on testing EHR data, utilizing various evaluation metrics.

%
%
%
%
%
%

%% file: Experiment.tex

\subsection{Datasets}

The N2C2 2022 competitions introduced the Contextualized Medication Event Dataset (CMED) specifically designed to capture the contextual information regarding medication changes within electronic health records (EHRs). CMED was created by a team of three annotators who annotated over 500 randomly selected EHRs. The dataset encompasses 9,013 medication mentions, each carefully labeled based on its context in the EHRs. The labels assigned to each medication mention include ``Disposition" for discussing a change, ``NoDisposition" for no change, and ``Undetermined" when further information is required to determine the change for medication event classification.

\begin{figure} [ht]
 	\centering	\includegraphics[scale=0.5]{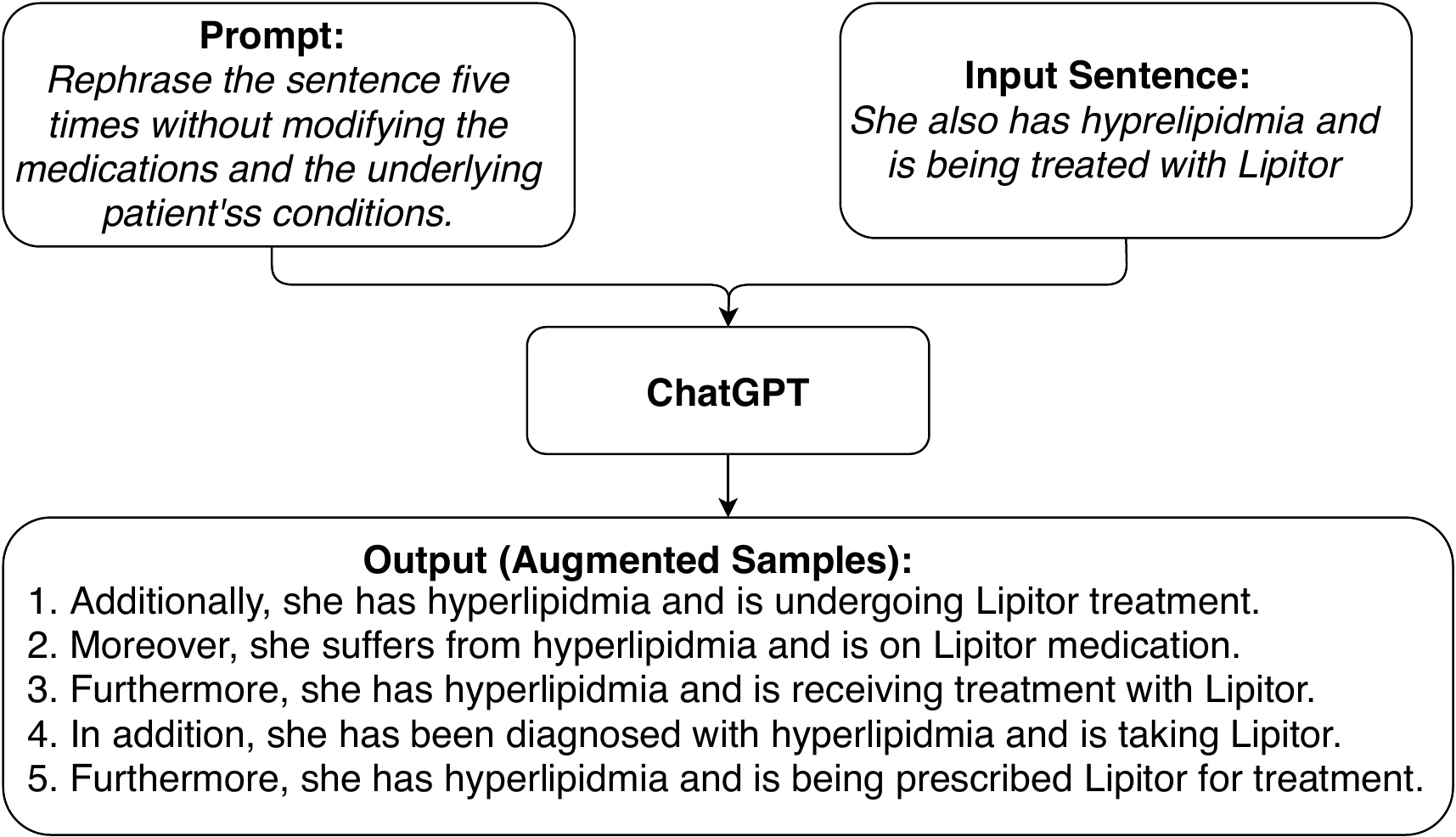}
	\caption{An example of generated samples via ChatGPT}
	\label{Fig_4_1}
\end{figure}

In addition, data augmentation using ChatGPT is employed to enhance the diversity and volume of the training data for medication identification and event classification. Figure~\ref{Fig_4_1} illustrates an example of the generated samples through ChatGPT. These samples retain the essential information concerning the patient's condition (hyperlipidemia) and the medication (Lipitor), while presenting them in various syntactic structures. A 10\% portion of the training data is utilized for augmentation in this study.

\subsection{Evaluation Metrics}
\label{sec:metrics}


For evaluating the performance of our proposed data augmentation in medication event classification, we treated it as a multiclass classification problem. To assess the effectiveness, we utilized various evaluation metrics, including macro-average Precision (MacroP), macro-average Recall (MacroR), macro-average F-score (MacroF)~\cite{van2013macro}, as well as micro-average Precision (MicroP), micro-average Recall (MicroR), and micro-average F-score (MicroF)~\cite{van2013macro}. Moreover, Macro-average involves calculating Precision, Recall, and F-scores independently for each type of medication event and then averaging these metrics. It provides an overall assessment of the medication event classification performance~\cite{yang2001study}. Additionally, we considered both strict and lenient matching criteria in calculating these evaluation metrics (multi). Strict matching indicates that the two offsets associated with the medication must match exactly, while lenient matching allows for an overlap between the two offset pairs.

%
%
%
%

\subsection{Pre-trained BERT Models}
Table~\ref{tab:pretraining} shows the variants of Pretrained BERT models employed in this paper.\\
\begin{table}[ht]
\caption{Variants of Pretrained BERT models. The main difference between $Roberta_{base}$  and $BERT_{base}$ is that it applied different hyper-parameters to train BERT models.}
\centering
\begin{tabular}{|l|l|}

\hline
\textbf{Model} & \textbf{Pre-training Data}\\
\hline

$BERT_{base}$~\cite{bert} & Wikipedia and Bookscorpus	 \\ 
$BioBERT_{pmc}$~\cite{biobert} & PMC Journals\\
$BioBERT_{pubmed}$ & PubMed Abstracts \\
$BioBERT_{pubmed\_pmc}$	&PubMed Abstracts, PMC Journals \\
$Clinical BERT$~\cite{cbert}	&MIMIC Notes \\
$Discharge BERT$  & MIMIC-III Discharge Notes \\
$BioClinical BERT$ 	 &PubMed Abstracts, PMC Journals, MIMIC \\
$BioDischarge BERT$ & PubMed Abstracts, PMC Journals, MIMIC-III  \\
$BioReddit BERT$~\cite{reddit}	& PubMed Abstracts, PMC Journals, Reddit \\
$Roberta_{base}$~\cite{roberta} & Wikipedia and Bookscorpus \\
$Roberta_{large}$ & Wikipedia and Bookscorpus \\

\hline

\end{tabular}
\label{tab:pretraining}
 \end{table}

\subsection{Result and Discussion}


\subsubsection{Medication Identification}

The medication identification task in this study is treated as a binary classification problem, with two classes: ``drug" and ``non-drug." Table~\ref{table-med} presents a performance comparison between pre-trained BERT models trained on the original data and augmented training data for medication identification. It is observed that the pre-trained BERT models demonstrate promising performance based on the evaluation results. It is worth noting that BERT models pre-trained on biomedical data, such as BioBERT, achieve higher performance. This suggests that incorporating biomedical data in the pre-training process contributes to improved performance in medication identification.


Moreover, regarding the proposed data augmentation by ChatGPT, Enhanced $Roberta_{large}$, Roberta trained on the augmented dataset plus training data, shows the optimal performance, which means that training models on an augmented dataset can further improve the performance. Based on the observations mentioned earlier, the proposed data augmentation technique using ChatGPT can enhance the performance of medication identification. By augmenting the training data with additional examples generated by ChatGPT, the models are exposed to a wider range of variations and contexts related to medication events. This increased diversity in the training data helps the model better capture the patterns associated with medication events in clinical notes.

\begin{table*}
\caption{Performance comparison on medication identification. Enhanced $Roberta_{large}$ is built with data augmentation.}
\centering
\begin{tabular}{|l|ccc|ccc|}
\hline
\textbf{Model}&\multicolumn{3}{c|}{\textbf{Strict Evaluation Performance}}&\multicolumn{3}{c|}{\textbf{Lenient Evaluation Performance}} \\
\hline
\textbf{Pretrained BERT}  & \textbf{Precision} & \textbf{Recall} & \textbf{Fscore} & \textbf{Precision} & \textbf{Recall} & \textbf{Fscore} \\
\hline
$BERT_{base}$	&0.8493  & 0.7879 &0.8181 &	 0.8791 &0.8169	&0.8469\\
$BioBERT_{pmc}$	&0.8521	&0.7971	&0.8237 & \textbf{0.8830}		&0.8260	&0.8535\\
$BioBERT_{pubmed}$&	0.8508&	0.8084	&0.8291 & 0.8759	&0.8322&	0.8585\\
$BioBERT_{pubmed\_pmc}$	&\textbf{0.8669}	&0.7965&	0.8309 & 0.8813	&0.8388	&0.8552\\
$Clinical BERT$	&0.8421	&0.8135	&0.8276 &0.8685	&0.8390	&0.8535\\
$Discharge BERT$	&0.8507	&0.8082	&0.8287 &0.8788	&0.8345	&0.8561\\
$BioClinical BERT$	&0.8525&	0.8229	&0.8373 &0.8702&	0.8396	&0.8546\\
$BioDischarge BERT$	&0.8381	&0.8201	&0.8241 &0.8710&	0.8418	&0.8572\\
$BioReddit BERT$	&0.8379	&\textbf{0.9256}	&0.8802 &0.8670&	\textbf{0.9609}	&0.9115\\
$Roberta_{base}$	&0.8382	&0.9233	&0.8783 &0.8650	&0.9518	&0.9063\\
$Roberta_{large}$	&0.8484	&0.9235&	0.8844 &0.8724	&0.9495	&\textbf{0.9193}\\
\hline
\hline
\textbf{Enhanced $Roberta_{large}$}	&0.8660	&0.9190	&\textbf{0.8946}&0.8758	&0.9508&	\textbf{0.9224}\\

\hline
\end{tabular}
\label{table-med}
\end{table*}

\begin{table*}
\caption{Performance comparison on medication event classification via Micro-average evaluation metrics.}
\centering
\begin{tabular}{|l|ccc|ccc|}
\hline
\textbf{Model}&\multicolumn{3}{c|}{\textbf{Strict Evaluation Performance}}&\multicolumn{3}{c|}{\textbf{Lenient Evaluation Performance}} \\
\hline
\textbf{Pretrained BERT}  & \textbf{MicroP} & \textbf{MicroR} & \textbf{MicroF} & \textbf{MicroP} & \textbf{MicroR} & \textbf{MicroF} \\
\hline
$BERT_{base}$	&0.7297&	0.6776&	0.7027 &	 0.7572&	0.7031&	0.7291\\
$BioBERT_{pmc}$	&0.7382	&0.6901	&0.7133 & 0.7670	&0.7170 &0.7420\\
$BioBERT_{pubmed}$&	0.7399	&0.7072	&0.7207 & 0.7637&	0.7252	&0.7440\\
$BioBERT_{pubmed\_pmc}$	&0.7381	&0.6944&	0.7156 & 0.7619&	0.7271	&0.7483\\
$Clinical BERT$	&0.7453&	0.7199&	0.7322 &0.7682	&0.7416	&0.7547\\
$Discharge BERT$	&0.7369	&0.6986	&0.7169 &0.7642	&0.7252	&0.7442\\
$BioClinical BERT$	&0.7396	&0.7188	&0.7317 &0.7620	&0.7348	&0.7482\\
$BioDischarge BERT$	&0.7296	&0.7048	&0.7170 &0.7607&	0.7358&	0.7476\\
$BioReddit BERT$	&0.7556&	0.8238	&0.7817 &0.7758	&0.8567	&0.8129\\
$Roberta_{base}$	&0.7461	&0.8210	&0.7818 &0.7719	&0.8493	&0.8087\\
$Roberta_{large}$	&0.7678	&0.8357&	0.8003 &0.7902	&0.8601	&0.8237\\
\hline
\hline
\textbf{Enhanced $Roberta_{large}$ }	&\textbf{0.7698}	&\textbf{0.8392}	&\textbf{0.8104}&\textbf{0.7985}	&\textbf{0.8612}&	\textbf{0.8268}\\
\hline
\end{tabular}
\label{tab-med-micro}
\end{table*}

\begin{table*}
\caption{Performance comparison on medication event classification via Macro-average evaluation metrics.}
\centering
\begin{tabular}{|l|ccc|ccc|}
\hline
\textbf{Model}&\multicolumn{3}{c|}{\textbf{Strict Evaluation Performance}}&\multicolumn{3}{c|}{\textbf{Lenient Evaluation Performance}} \\
\hline
\textbf{Pretrained BERT}  & \textbf{MacroP} & \textbf{MacroR} & \textbf{MacroF} & \textbf{MacroP} & \textbf{MacroR} & \textbf{MacroF} \\
\hline
$BERT_{base}$	&0.6344	&0.5960&	0.6145 &	 0.6491&	0.6097&	0.6287\\
$BioBERT_{pmc}$	&0.6835& 0.6011&	0.6349 & 0.6989	&0.6146&	0.6500\\
$BioBERT_{pubmed}$&	0.6559&	0.6348&	0.6436 & 0.6696	&0.6471&0.6570\\
$BioBERT_{pubmed\_pmc}$	&0.6549	&0.6073	&0.6298 & 0.6723	&0.6237&	0.6467\\
$Clinical BERT$	&0.6830	&0.6887&	0.6843 &0.6943	&0.6993&	0.6956\\
$Discharge BERT$	&0.6633&	0.6479&	0.6547 &0.6775	&0.6613	&0.6685\\
$BioClinical BERT$	&0.6596	&0.6812	&0.6687 &0.6750	&0.6960	&0.6838\\
$BioDischarge BERT$	&0.6540	&0.6703&	0.6610 &0.6725	&0.6884&	0.6793\\
$BioReddit BERT$	&0.6774	&0.6911	&0.6816 &0.6925	&0.7096&	0.6976\\
$Roberta_{base}$	&0.6737	&0.7235	&0.6975 &0.6880	&0.7393	&0.7126\\
$Roberta_{large}$	&0.7114	&\textbf{0.7274}&	0.7176 &0.7250	&0.7423	&0.7317\\

\hline
\hline
\textbf{Enhanced $Roberta_{large}$ }	&\textbf{0.7124}	&0.7168 &\textbf{0.7198}	&\textbf{0.7287}&\textbf{0.7426}	&\textbf{0.7356}\\

\hline
\end{tabular}
\label{tab-med-macro}
\end{table*}

\subsubsection{Medication Event Classification}

Table~\ref{tab-med-micro} and~\ref{tab-med-macro} shows the performance comparison on medication events classification via Micro-average and Macro-average evaluation metrics. Similar to medication identification, pre-trained BERT models are able to obtain promising performance based on strict and lenient evaluation results. For example, in table~\ref{tab-med-micro}, $Roberta_{large}$ could obtain 0.8003 and 0.8237 for strict and lenient F-score, respectively. Enhanced  $Roberta_{large}$  was able to achieve 0.8104 and 0.8268 for strict and lenient F-score. For Macro-average evaluation results in Table~\ref{tab-med-macro}, the proposed data augmentation enhanced the performance on lenient evaluation metrics. 

By comparing the results obtained with and without data augmentation, it becomes evident that the augmentation technique positively impacts the performance of medication events classification. The utilization of data augmentation by ChatGPT serves as a valuable approach to enhance the performance of medication event classification tasks, providing more accurate and reliable results.

%% file: Relatedwork.tex
The unstructured nature of EHRs data poses challenges when applying machine learning algorithms to analyze it efficiently. Previous research has indicated that conventional deep learning models such as Recurrent Neural Networks (RNN) and Long Short-term Memory (LSTM) perform poorly due to the infrequent appearance of medical and biomedical terms in spoken and written language~\cite{bilstm}. Additionally, using deep learning models as transfer learning on EHR datasets alone yields limited performance due to the scarcity of biomedical and medical terms in general language~\cite{bert}. To overcome these limitations, pre-training deep learning models with publicly available medical and biomedical data has been shown to improve performance on EHR data~\cite{biobert, elmo, cbert}.

Generative pre-training has proven highly effective in addressing challenging natural language processing (NLP) tasks that conventional NLP systems struggle with~\cite{4}. Recent studies have explored the use of ChatGPT for data augmentation, leveraging its ability to generate diverse data. By rephrasing sentences and providing alternative phrasings, ChatGPT enhances the deep learning model's capability to comprehend and capture more intricate features within the data~\cite{3}. The introduction of data augmentation techniques that harness the power of large-scale language models to generate synthetic data closely mimicking the patterns and characteristics of the original training data has shown promise~\cite{2}.


%% file: Conclusion.tex
This paper explores the application of ChatGPT for data augmentation in EHR data analysis. It focuses on two tasks: medication identification and medication event classification from the N2C2 2022 competitions. Multiple variants of pre-trained BERT models are employed in the study. The experimental results, evaluated using strict and lenient metrics encompassing precision, recall, and F-score, emphasize the performance of pre-trained BERT models in these tasks. Particularly, the study highlights the advantages of models pre-trained on biomedical data and augmented with examples generated by ChatGPT. In essence, the findings indicate that data augmentation using ChatGPT proves advantageous in enhancing the performance of both medication identification and medication event classification.